\documentclass{aims}
\usepackage{amsmath}
\usepackage{mathtools}
\usepackage{makecell}
\usepackage{amssymb}
\usepackage[ruled,vlined,linesnumbered]{algorithm2e}
\usepackage{paralist}
\usepackage{graphics} 
\usepackage{epsfig} 
\usepackage[colorlinks=true]{hyperref}
\hypersetup{urlcolor=blue, citecolor=red}

\def\currentvolume{X}
 \def\currentissue{X}
  \def\currentyear{200X}
   \def\currentmonth{XX}
    \def\ppages{X--XX}

\theoremstyle{definition}

\DeclareMathOperator*{\argmin}{arg\,min}

\def \Re {\mathbb{R}}

\title[Surveillance Video Processing]
      {Surveillance Video Processing Using Compressive Sensing}

\author[Hong Jiang, Wei Deng and Zuowei Shen]{}


 \email{hong.jiang@alcatel-lucent.com}
 \email{wei.deng@rice.edu}
 \email{matzuows@nus.edu.sg}

\thanks{}

\begin{document}
\maketitle

\centerline{\scshape Hong Jiang}
\medskip
{\footnotesize
 \centerline{Bell Labs, Alcatel-Lucent}
   \centerline{700 Mountain Ave}
   \centerline{Murray Hill, NJ 07974, USA}
} 

\medskip

\centerline{\scshape Wei Deng}
\medskip
{\footnotesize
 \centerline{Dept. of Computational and Applied Math}
   \centerline{Rice University}
   \centerline{Houston, TX 77005, USA}
}

\medskip

\centerline{\scshape Zuowei Shen}
\medskip
{\footnotesize
 \centerline{Dept. of Math}
   \centerline{National Univ. of Singapore}
   \centerline{Singapore 119076}
}

\bigskip

 \centerline{(Communicated by the associate editor name)}

\begin{abstract}
A compressive sensing method combined with decomposition of a matrix formed with image frames of a surveillance video into low rank and sparse matrices is proposed to segment the background and extract moving objects in a surveillance video. The video is acquired by compressive measurements, and the measurements are used to reconstruct the video by a low rank and sparse decomposition of matrix. The low rank component represents the background, and the sparse component is used to identify moving objects in the surveillance video. The decomposition is performed by an augmented Lagrangian alternating direction method. Experiments are carried out to demonstrate that moving objects can be reliably extracted with a small amount of measurements.
\end{abstract}

\section{Introduction}

In a network of cameras for surveillance, a massive number of cameras are deployed, some with wireless connections. The cameras transmit surveillance videos to a processing center where the videos are processed and analyzed. Of particular interest in surveillance video processing is the ability to detect anomalies and moving objects in a scene automatically and quickly.

Detection of moving objects is traditionally achieved by background subtraction methods \cite{Benezeth, Piccardi} which segment background and moving objects in a sequence of surveillance video frames. The mixture of Gaussians \cite{Stauffer} technique assumes that each pixel has a distribution that is a sum of Gaussians and the background and foreground are modeled by the size of the Gaussians. In low rank and sparse decomposition \cite{Candes2009}, the background is modeled by a low rank matrix, and the moving objects are identified by a sparse component. These traditional background subtraction techniques require all pixels of a surveillance video to be captured, transmitted and analyzed.

A challenge in the network of cameras is the bandwidth. Since traditional background subtraction requires all pixels of video to be acquired, an enormous amount of data is transported in the network due to a large number of cameras. At the same time, most of the data is uninteresting due to inactivity. There is a high risk of the network being overwhelmed by the mostly uninteresting data to prevent timely detection of anomalies and moving objects. Therefore, it is highly desirable to have a network of cameras in which each camera transmits a small amount of data with enough information for reliable detection and tracking of moving objects or anomalies. Compressive sensing \cite{Candes2005, Donoho} allows us to achieve this goal. In compressive sensing, the surveillance cameras make compressive measurements of video and transmit measurements in the network. Since the number of measurements is much smaller than the total number of pixels, transmission of measurements, instead of pixels, helps to prevent network congestion. Furthermore, the lower data rate of compressed measurements helps wireless cameras to reduce power consumption.

When a surveillance video is acquired by compressive measurements, the pixel values of the video frames are unknown, and consequently, the traditional background subtraction techniques such as \cite{Candes2009, Stauffer} cannot be applied directly. A straight forward approach is to recover the video from the compressive measurements \cite{Jiang, Li}, and then, after the pixel values are estimated, to apply one of the known background subtraction techniques. Such an approach is undesirable for two reasons. First, a generic video reconstruction algorithm does not take advantage of special characteristics of surveillance video in which a well defined, relatively static background exists. The existence of a background provides prior information that helps to reduce the number of measurements. Secondly, in the straight forward approach, additional processing is needed to perform background subtraction after the video is recovered from the measurements.

In this paper, we propose a method for segmentation of background by using a low rank and sparse decomposition of matrix. In this method, the compressive measurements from a surveillance camera are used to reconstruct video which is assumed to be comprised of a low rank and a sparse component. As in \cite{Candes2009}, the low rank component is the background, and the sparse component identifies moving objects. Therefore, the background subtraction becomes part of the reconstruction, and no additional processing is needed after reconstruction. Furthermore, the reconstruction takes advantage of the knowledge that there exits a background in the video, which helps to reduce the number of measurements required.

The proposed method is inspired by the work of \cite{Candes2009} and extends it to the measurement domain, rather than the pixel domain, for use in conjunction with compressive sensing. This method is motivated by \cite{Gao} where a matrix equation is solved with the assumption that the solution is a sum of a low rank matrix and a sparse matrix for 4D-CT reconstruction. Compressive sensing has been used in background subtraction previously \cite{Cevher}, but the method of \cite{Cevher} requires the pixel values of the background to be known a priori, such as acquired from a training process. The method of this paper may be considered to be the training process in which the compressive measurements are used to obtain the background.

The paper is organized as follows. In Section 2, the framework for reconstruction by low rank and sparse decomposition is introduced. The alternative direction method (ADM) for solving the optimization problem is discussed in Section 3. The treatment of color components is discussed in Section 4. Finally, experiments are discussed and results are reported in Section 5.

\section{Low rank and sparse decomposition}

\begin{figure}[htp]
\begin{center}
  \includegraphics[width=4in]{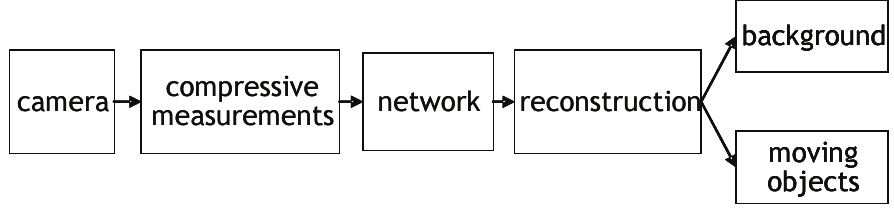}\\
  \caption{Compressive video sensing framework}\label{fig1}
  \end{center}
\end{figure}

The framework of our method is shown in Figure \ref{fig1}. We first treat the video as black and white, having only the luminance component. Color video with R, G, B components will be discussed later.

\subsection{Video volume}
We consider a video sequence consisting of a number of frames. Let $x_j \in \Re^n$ be a vector formed from pixels of frame $j$ of the video sequence, for $j = 1, \ldots, J$, where $J$ is the total number of frames and $n$ is the total number of pixels in a frame. Let $X = [x_1,\ldots,x_J] \in \Re^{n\times J}$ be the matrix of dimension $n \times J$, the columns of which are the frames in the video sequence.

In general, $X=[x_1,\ldots,x_J] \in \Re^{n \times J}$ is a video volume obtained from a video sequence, in which each $x_j$ is a vector formed from pixels of a sub-region in frame $j$ of  the video sequence. The position of the sub-region within each frame is independent of $j$. The index $n$ is the total number of pixels in the sub-region. The total number of entries in $X$  is $N=nJ$.

\subsection{Compressive measurements}
Let $\phi$ be an $M \times N$ measurement matrix with $M$ rows and $N$ columns, where $M<N$. The measurement matrix $\phi$ may be chosen as a random matrix such as a randomly permutated Walsh-Hadamard matrix. Let $\phi=[\phi_1,\ldots,\phi_J]$ , where $\phi_j \in \Re^{M \times n}$  is a matrix of dimension $M \times n$

The compressive measurements of the video volume are defined as
  \begin{equation}\label{eq1}
       y= \phi \circ X \triangleq \sum_{j=1}^{J} \phi_j x_j,
  \end{equation}
where $y$ is a vector of length $M$. The number of measurements, $M$, is much smaller than the total number of pixels of the video volume, $N$. The rest of the processing will only make use of the measurements $y$, without knowing the original video volume $X$.

The process of making compressive measurements may be considered to be an encoding of the video volume, in which the video volume is encoded by compressive measurements. In compressive sensing, the encoding is theoretically a matrix-by-vector multiplication. How well a video $X$ can be recovered from compressive measurements $y$ depends on the sparsity of the $X$ (after transform) and the properties of the measurement matrix $\phi$.  It is well known that if $\phi$ satisfies the restricted isometry property (RIP), then the signal $X$ can be recovered from the measurements $y$ if the number of measurements $M$ is large enough \cite{Candes2005, Donoho}. Randomly permutated Walsh-Hadamard matrices are shown to have RIP \cite{Rudelson}, and such matrices have been successfully used as measurement matrices in compressive video sensing \cite{Jiang, Li}. 

Although the measurements are defined by a matrix multiplication, the operation of matrix-by-vector multiplication is seldom used in practice, because it has a complexity of $\mathcal{O}(MN)$ which may be too expensive for real time applications. When a randomly permutated Walsh-Hadamard matrix is used as the sensing matrix, the measurements may be computed by using a fast transform which has complexity of $\mathcal{O}(N \log(N))$\cite{Sutter}. Acquisition of measurements by using other sensing matrices, such as a circulant matrix generated by a pseudo-random sequence \cite{Jiang3}, can also be implemented very efficiently in hardware by using shift registers.

\subsection{Reconstruction}
Given the measurement vector $y$, the video volume $X$ can be reconstructed by using the following minimization problem:

  \begin{equation}\label{eq2}
       X = X_1 + X_2,
  \end{equation}

  \begin{equation}\label{eq3}
  \begin{split}
       (X_1,X_2) = &\argmin_{X_1, X_2}\, \mu_1 ||X_1||_* + \mu_2 ||W_s^1 X_1||_1 +\mu_3 ||W_s^2 X_2||,\\
                   &\text{s.t. } y = \phi \circ X.
  \end{split}
  \end{equation}

In \eqref{eq3}, $||A||_*$ is the nuclear norm of a matrix $A \in \Re^{n \times J}$ defined by

  \begin{equation}\label{eq4}
       ||A||_* \triangleq \text{trace} (\sqrt{A^*A}) = \sum_{i=1}^{\text{min}(n,J)} \sigma_i,
  \end{equation}
where $\sigma_{i}$ are the singular values of matrix $A$. The nuclear norm of $A$ is the $l_1$-norm of its singular values. $||A||_1$ is the $l_1$-norm when $A$ is considered to be a vector, i.e., $||A||_1 \triangleq \sum_{i=1}^n \sum_{j=1}^{J} |a_{ij}|$. $\mu_1$, $\mu_2$ and $\mu_3$ are some nonnegative constants. $W_s^i$, $i = 1,2$ are transforms that give sparse representations of underlying frames or video.  The transform used in this paper is the wavelet frame transform constructed in \cite{DongB} which will be described later.

In  \eqref{eq2}, $X_1$ and $X_2$ represent two different components of the reconstructed video volume. The low rank component $X_1$ is a relatively stationary component, which represents the background of the video. For example, if $X_1$ has rank one, then $X_1 = [c_1 x_b, \ldots, c_J x_b] \in \Re^{n \times J}$ , where $x_b$ is the vector formed from pixels of the background image, and $c_j$ are some constants. In other words, $X_1$ is made up of a sequence of the still images which are scaled images of the stationary background in the video. Matrix $X_2$ of \eqref{eq2} is the sparse component, which represents moving objects in the video volume.

\subsection{Sparsifying operators}
The background image $x_b$ may be sparse in some transformed space, for example, in a wavelet transform space. Similarly, the moving objects represented by $X_2$ may have spatial correlations which can be sparsified by a transform.

The operators $W_s^i$, $i=1,2$ in \eqref{eq3} are sparsifying, spatial operators. Because $W_s^i$, $i=1,2$ have the same form, for simplicity, we use $W_s$ to denote each of $W_s^i$, $i=1,2$. For a given matrix $A \in \Re^{n \times J}$, the operators $W_s$ work on columns of the matrix $A$. Specifically, let
  \begin{equation}\label{eq5}
       A = [a_1, \ldots, a_J], \;\; a_j \in \Re^{n}.
  \end{equation}
Then the spatial operator   is defined as
  \begin{equation}\label{eq6}
       W_s A \triangleq [W_1 a_1, \ldots, W_J a_J], W_j \in \Re^{n' \times n}, j = 1, \ldots, J.
  \end{equation}
In other words, the spatial operator $W_s$ is defined by $J$ linear operators that can be generated by wavelet decomposition algorithm on each image frame as given by \cite{Daubechies}. It can be represented by matrices $W_j$, $j=1, \ldots, J$ of dimension $n' \times n$, often  $n' \geq n$. The operator $W_j$, $j=1,\ldots,J$ may be different for each of $W_s^i$, $i=1,2$ but they also may be the same. Furthermore, the matrices $W_j$ may be identical, i.e., $W_j = W_0$, for all $j=1,\ldots, J$. The matrices $W_j$ are chosen to be the tight frame transform as given in \cite{Daubechies, Ron}. The wavelet frames are used in image restorations, since they give sparse approximations for many images. More details on applications of wavelet frame for image restorations can be found in \cite{DongB, Shen}.


\section{Minimization Algorithm}
The minimization problem \eqref{eq3} is a convex problem, so standard convex optimization algorithms such as the interior point methods can be used. However, these standard methods are computationally expensive and may not have the required sparsity and low rank of the solution when the approximated minimizer is derived from them. Instead, as shown in \cite{Cai2010}, the singular value threshold method is very efficient in low rank matrix completion and low rank matrix and sparse matrix decomposition. We use the idea of singular value threshold based first-order method using the Augmented Lagrangian Alternating Direction (ADM) for solving this minimization problem. We remark that this method is similar to the split Bregman method used in image restorations, see \cite{Cai2009, Goldstein} for details.

We first reformulate the problem \eqref{eq3} into an equivalent problem by introducing some splitting variables as follows:
  \begin{equation}\label{eq7}
  \begin{split}
        \min_{X_1,X_2} &\,\mu_1 ||Z_1||_* + \mu_2 ||Z_2||_1 + \mu_3 ||Z_3||_1,\\
        \text{s.t. } &X_1 = Z_1, \; W_s^1 X_1 = Z_2, \; W_s^2 X_2 = Z_3,\\
       & \phi \circ (X_1 + X_2) = y.
  \end{split}
  \end{equation}
We solve this problem by applying the ADM framework, an iterative procedure that minimizes the augmented Lagrangian function  in alternating directions and updates the Lagrangian multipliers in every iteration. The benefit of the alternating minimization approach is that it divides the original problem into some subproblems which either have closed form solutions or can be solved efficiently.

Specifically, the augmented Lagrangian of problem \eqref{eq7} is given by
  \begin{equation}\label{eq8}
  \begin{split}
  \mathcal{L}_A = &\,\mu_1 ||Z_1||_* + \mu_2 ||Z_2||_1 + \mu_3 ||Z_3||_1 \\
      & - \langle \lambda_1, X_1 - Z_1 \rangle + \dfrac{\beta_1}{2} ||X_1 -Z_1||_F^2 \\
      & - \langle \lambda_2, W_s^1 X_1 - Z_2 \rangle + \dfrac{\beta_2}{2} ||W_s^1 X_1 -Z_2||_F^2 \\
      & - \langle \lambda_3, W_s^2 X_1 - Z_3 \rangle + \dfrac{\beta_3}{2} ||W_s^2 X_1 -Z_3||_F^2  \\
      & - \langle \lambda_4, \phi \circ (X_1+X_2)-y \rangle +\dfrac{\beta_4}{2} ||\phi \circ (X_1+X_2)-y||_F^2,
  \end{split}
  \end{equation}
where $\lambda_i$ $(i=1, \ldots,4)$ are Lagrangian multipliers, and $\beta_i >0$ $(i = 1,\ldots,4)$ are penalty parameters.

In each iteration, the augmented Lagrangian is minimized over $X$- and $Z$- directions alternately and the Lagrangian multipliers are updated by the following simple scheme:
  \begin{equation}\label{eq9}
  \begin{cases}
  & \lambda_1 \leftarrow \lambda_1 - \gamma \beta_1 (X_1 - Z_1) \\
  & \lambda_2 \leftarrow \lambda_2 - \gamma \beta_2 (W_s^1 X_1 -Z_2)  \\
  & \lambda_3 \leftarrow \lambda_3 - \gamma \beta_3 (W_s^2 X_2 -Z_3) \\
  & \lambda_4 \leftarrow \lambda_4 - \gamma \beta_4 [\phi \circ (X_1+X_2) -y]
  \end{cases},
  \end{equation}
where $\gamma >0$ is a step-length.

Clearly, the $(X_1,X_2)$-subproblem, i.e., to minimize \eqref{eq8} over $(X_1,X_2)$ is a convex quadratic problem, which reduces to solving a linear system $\nabla_{(X_1,X_2)} \mathcal{L}_A =0$. When the sparsifying transforms $W_s^i$ form tight frames, i.e., $W_s^{iT} W_s^i = I$ $(i=1,2)$ and the rows of the measurement matrix are orthonormal, i.e., $\phi \phi^T = I$, the linear system can be solved  by Schur complement and Sherman-Morrison-Woodbury formula, in which the major computations are only matrix-vector multiplications without inversion of a linear system. In other cases, solving a linear system may be too expensive for large scale data. However, the linear system can be solved approximately, e.g., by just taking a steepest descent step, and empirical evidence shows that the convergence of the algorithm can still be well achieved.

Note the variables $Z_1,Z_2,Z_3$ are separable in the augmented Lagrangian function $\mathcal{L}_A$. Therefore, minimizing $\mathcal{L}_A$ over $(Z_1,Z_2,Z_3)$ boils down to minimizing over each $Z_i$ $(i=1,2,3)$ separately, i.e.,
\begin{align}
Z_1 =& \argmin_{Z_1} \;\; \mu_1 ||Z_1||_* + \dfrac{\beta_1}{2} ||Z_1 - \left(X_1 - \lambda_1/\beta_1 \right)||_F^2, \label{eq10} \\
Z_2 =& \argmin_{Z_2} \;\; \mu_2 ||Z_2||_1 + \dfrac{\beta_2}{2} ||Z_2 - \left(W_s^1 X_1 - \lambda_2/\beta_2 \right)||_2^2, \label{eq11a}\\
Z_3 =& \argmin_{Z_3} \;\; \mu_3 ||Z_3||_1 + \dfrac{\beta_3}{2} ||Z_3 - \left(W_s^2 X_2 - \lambda_3/\beta_3 \right)||_2^2. \label{eq11b}
\end{align}
All of them are known to have closed form solutions. The subproblem \eqref{eq10} can be solved by so-called singular value thresholding (SVT), i.e.,
\begin{equation}\label{eq12}
Z_1 = D_{\mu_1/\beta_1} (X_1 -\lambda_1/\beta_1),
\end{equation}
where $D_\tau (\cdot)$ denotes the SVT operator as follows:
\begin{equation}\label{eq13}
D_\tau (X) \triangleq U \cdot \text{diag} (\text{max} (\sigma - \tau, 0)) \cdot V^T,
\end{equation}
and $X= U \cdot \text{diag}(\sigma) \cdot V^T$ is the singular value decomposition (SVD) of the input matrix $X$. The subproblems \eqref{eq11a} and \eqref{eq11b} can be solved by the so-called shrinkage formula. Let
\begin{equation}\label{eq14}
T_\tau (x) \triangleq \text{sgn}(x) \cdot \text{max} (|x| - \tau,0),
\end{equation}
denote the shrinkage operator, then the solutions to \eqref{eq11a} and \eqref{eq11b} are given by
\begin{align}
Z_2 =& T_{\mu_2/\beta_2} (W_s^1 X_1 - \lambda_2/\beta_2),\\
Z_3 =& T_{\mu_3/\beta_3} (W_s^2 X_2 - \lambda_3/\beta_3). \label{eq15}
\end{align}
The iterative scheme of the algorithm is summarized as below.

\begin{algorithm}[H]
\SetKwInOut{Input}{input}\SetKwInOut{Output}{output}
\SetKwComment{Comment}{}{}
\BlankLine
Initialize $X_1,X_2,Z_1,Z_2,Z_3,\lambda_i,\beta_i$ $(i=1,\ldots,4)$ and $\gamma$\;
\While{stopping criterion is not met}{
compute $(X_1,X_2)$ from $\nabla_{X_1,X_2} \mathcal{L}_A =0$\;
$Z_1 = D_{\mu_1/ \beta_1} (X_1 -\lambda_1/ \beta_1)$\;
$Z_2 = T_{\mu_2/ \beta_2} (W_s^1 X_1 -\lambda_2/ \beta_2)$\;
$Z_3 = T_{\mu_3/ \beta_3} (W_s^2 X_2 -\lambda_3/ \beta_3)$\;
update $\lambda_i$ $(i=1,\ldots,4)$ by \eqref{eq9}\;}
\caption{ADM for Low-rank and Sparse Decomposition}
\end{algorithm}

Following from existing ADM theory, the algorithm has global convergence if $\beta_i > 0\,(i=1,\ldots,4)$ and $0 < \gamma < (\sqrt{5} +1) /2$, see \cite{Glowinski}.

\section{Color components}

So far, we have only considered the luminance component of a video volume, treated as a single matrix or vector. A color video has multiple color components, such as RGB components, corresponding to multiple matrices or vectors. Although each color component can be dealt with individually using our previous model, this approach certainly does not exploit the high correlations between different color components. Therefore, we want to take advantage of these correlations between color components and develop a joint reconstruction procedure. Compressive sensing using joint sparsity has previously been considered in, such as, \cite{Deng, Fornasier}.
\subsection{Correlations}
Let matrices $X_i^{(1)}$, $X_i^{(2)}$, $X_i^{(3)} \in \Re^{n \times J},$ $i=1,2$ denote the R, G, B components of low rank $(i=1)$ and sparse $(i=2)$ component, respectively. We then define a joint matrix of the colored video for each of low rank and sparse component by
\begin{equation}\label{eq16}
X_i = [X_i^{(1)T},X_i^{(2)T},X_i^{(3)T}]^T \in \Re^{3n \times J} \;, i=1,2.
\end{equation}

The correlations between color components can be considered in the following aspects. First, the linear dependency of background is often similar for different color components. As we mentioned, each $X_1^{(i)}$ $(i=1,2,3)$ tends to be low rank, and their columns are linearly dependent in a similar way for different colors. That means, by stacking them into a big matrix, the rank of $X_1$ will remain almost as low as each $X_1^{(i)}$ $(i=1,2,3)$. Secondly, different color components are likely to have similar sparsity structure under some sparsifying basis. For instance, if we apply wavelet frame transform to each color component, the large wavelet coefficients correspond to those locations with sharp changes of pixel values, i.e., the edges. However, note that the edges of an image are usually preserved across different color components. Therefore, under wavelet frame transforms, different color components become jointly sparse, sharing the same support locations. We therefore define the sparsifying operators on the joint components by
\begin{equation}
W^i X_i = [(W^i_s X_i^{(1)})^T, (W^i_s X_i^{(2)})^T, (W^i_s X_i^{(3)})^T]^T \in \Re^{3n' \times J} \;, i=1,2,
\end{equation}
where $n'$ and $W^i_s, i=1,2$ are defined in \eqref{eq6}.

\subsection{Joint reconstruction}

We extend our model \eqref{eq3} to deal with the joint reconstruction of multiple color components as below:
\begin{equation}\label{eq17}
\begin{split}
\min_{X_1,X_2} & \mu_1 ||X_1||_* + \mu_2 ||W^1 X_1||_{2,1} + \mu_3 ||W^2 X_2||_{2,1},\\
\text{s.t. }  &\phi \circ (X_1+X_2) = y.
\end{split}
\end{equation}
The mixed $\ell_{2,1}$-norm is defined to take into the consideration that now each pixel has three color components, or after sparsifying operator, each transformed coefficient is a 3-vector having three components (for R,G and B, respectively). In the mixed $\ell_{2,1}$-norm, therefore, the 2-norm of the 3-vector is computed first, and then 1-norm of the transformed coefficients is formed by
\begin{equation}\label{eq18}
||X||_{2,1} \triangleq \sum_{i=1}^{N'} \sqrt{\sum_{j=1}^3 (X^{(j)}(i))^2},
\end{equation}
for any $X=[X^{(1)T},X^{(2)T},X^{(3)T}]^T$. In \eqref{eq18}, $N'$ is total number of entries on each of $X^{(j)}$, $j=1,2,3$. Note that $N'$ may be different from $N$ because the sparsifying operator may be redundant. Since this $\ell_{2,1}$-norm is known to promote joint sparsity in the solution, we use it to encode the feature that color components have the same sparsity pattern under certain basis. And the nuclear norm $||\cdot||_*$ is applied to the joint matrix $X$ to exploit the correlations of background linear dependency for different color components.

\subsection{Algorithm}
The joint reconstruction model \eqref{eq17} can be solved efficiently by the ADM approach. Applying similar splitting technique, we transform \eqref{eq17} into an equivalent problem:
\begin{equation}
\begin{split}
\min_{X_1,X_2} &\mu_1 ||Z_1||_* + \mu_2 ||Z_2||_{2,1} + \mu_3 ||Z_3||_{2,1}, \\
\text{s.t. } & X_1 = Z_1,\; W^1 X_1 = Z_2,\; W^2 X_2 = Z_3, \\
& \phi \circ (X_1+X_2) = y.
\end{split}
\end{equation}
Following the same procedure, we derive an ADM algorithm as is summarized below.\\

\begin{algorithm}[H]
\SetKwInOut{Input}{input}\SetKwInOut{Output}{output}
\SetKwComment{Comment}{}{}
\BlankLine
Initialize $X_1,X_2,Z_1,Z_2,Z_3,\Lambda_i,\beta_i$ $(i=1,\ldots,4)$ and $\gamma$\;
\While{stopping criterion is not met}{
compute $(X_1,X_2)$ from $\nabla_{X_1,X_2} \mathcal{L}_A =0$\;
$Z_1 = D_{\mu_1/ \beta_1} (X_1 -\Lambda_1/ \beta_1)$\;
$Z_2 = S_{\mu_2/ \beta_2} (W^1 X_1 -\Lambda_2/ \beta_2)$\;
$Z_3 = S_{\mu_3/ \beta_3} (W^2 X_2 -\Lambda_3/ \beta_3)$\;
update $\Lambda_i$ $(i=1,\ldots,4)$:
\begin{equation*}
  \begin{cases}
  & \Lambda_1 \leftarrow \Lambda_1 - \gamma \beta_1 (X_1 - Z_1);  \\
  & \Lambda_2 \leftarrow \Lambda_2 - \gamma \beta_2 (W^1 X_1 -Z_2);  \\
  & \Lambda_3 \leftarrow \Lambda_3 - \gamma \beta_3 (W^2 X_2 -Z_3);  \\
  & \Lambda_4 \leftarrow \Lambda_4 - \gamma \beta_4 [\phi \circ (X_1+X_2) -y];
  \end{cases}
  \end{equation*}
}
\caption{ADM for Joint Low-rank and Joint Sparse Decomposition}
\end{algorithm}

Here, $S_\tau (\circ)$ represents a pixel-wise shrinkage operator, i.e.,
\begin{equation}
Z=S_\tau (X) \Leftrightarrow z_{ij} = \text{max} (x_{ij}-\tau, 0) \cdot \dfrac{x_{ij}}{||x_{ij}||_2}, \; \forall i,j,
\end{equation}
where $x_{ij}$ denotes the 3-vector from the input $X$ defined by $x_{ij}=[X_{ij}^{(1)},X_{ij}^{(2)},X_{ij}^{(3)}]^T$  and $z_{ij}$ is similarly defined. The same global convergence result follows \cite{Glowinski}.

\section{Numerical Experiment}

In this section, we present results from four numerical experiments.

\subsection{Experiment setup}
The surveillance video sequences, \emph{Browse2, ShopAssistant1Front, Traffic and Daniel\_light,} are obtained from databases that are publically available on the web \cite{DongY, Mahadevan, None}.

For each video sequence, a number of frames, ranging from 100 to 190 frames, are selected to form a video volume.
A permutated Walsh-Hadamard matrix is used to make measurements of the video volume.  The number of measurements used
in reconstruction of low rank and sparse decomposition is expressed in percentage of the total number of pixels in the
 video volume. For example, 100\% means the number of measurements is equal to the total number of pixels in the video volume.

In the reconstruction, the parameters $\mu_1,\mu_2,\mu_3$ are fixed for all four experiments, and they are given by
\begin{equation}
\mu_1 = 1,\; \mu_2=0,\; \mu_3= 1e-3.
\end{equation}

The parameter $\mu_2$ in \eqref{eq3} and \eqref{eq17} controls the amount of constraint imposed on the sparsity of the low rank component $X_1$. For the method of this paper, the constraint of low rank on $X_1$ is sufficient to produce a high quality reconstruction of the low rank component with a relatively small amount of measurements. The low rank component is common in most of the frames, and even though the percentage of measurement is small, there is a large amount of information about the low rank component $X_1$ in the measurements if the video volume has a large number of frames. For this reason, the parameter $\mu_2$ does not play an important role, and therefore, it is set to zero in the experiments of this paper. However, we introduce $\mu_2$ in \eqref{eq3} and \eqref{eq17} for a general framework, which can be also used in an adaptive method for real-time processing, see \cite{Jiang2}. In an adaptive method for real-time processing, the constraint of low rank alone is not sufficient to produce a high quality low rank component $X_1$ because the number of frames in the video is small. Therefore, the additional constraint of sparsity on $X_1$ becomes important. The effect of $\mu_2$ is discussed in detail in \cite{Jiang2}.

For each experiment, we report the PSNR of the reconstructed video, $X=X_1+X_2$. The experiments are summarized in the following table.
\begin{center}
\begin{table}[htp]
    \caption{Summary of experiments}
    \begin{tabular}{ | l | l | l | l | l |}
    \hline
    Name & Browse & Shop & Traffic & Daniel \\ \hline
    Resolution & $384 \times 288$ & $384 \times 258$ & $378 \times 282$ & $320 \times 240$
    \\ \hline
    Frames & 100 & 120 & 190 & 130
     \\ \hline
    Measurements (\%) & 4 & 4 & 6.67 & 10
    \\ \hline
    PSNR (dB) & 32.1 & 36.3 & 36.5 & 30.4
    \\ \hline
    Rank of $X_1$ & 1 & 1 & 1 & 3
    \\ \hline
    \end{tabular}
    \end{table}
\end{center}

To demonstrate the capability of detecting moving objects, we will display the images of the background $X_1$, and the silhouette of the moving objects obtained from the sparse component $X_2$. The silhouette of the $n$-th frame, $S_n$, is a binary image obtained from $X_2$ by the following equation.

\begin{equation}\label{eq22}
S_n = T_\delta (Med(X_2 (n)))).
\end{equation}

In \eqref{eq22}, $X_2 (n)$ is frame $n$ of the sparse component $X_2$. $Med(\cdot)$ a median filter, and $T_\delta (\cdot)$ is a threshold operator defined as
\begin{equation}
T_\delta(X)(i,j)=
\begin{cases}
1, & \text{if} \;|X(i,j)| \geq \delta  \\
0, & \text{if} \;|X(i,j)| < \delta
\end{cases}.
\end{equation}

\subsection{Browse2}
Browse2 \cite{None} is a color sequence from a camera monitoring a building lobby. The original is an MPEG file of resolution $384\times288$ and more than 6 minutes in length. We take 100 frames from the sequence, and process only the luminance component. The total number of pixels is $N=384\times288\times100 = 11059200$. The total number of measurements used in the reconstruction is 1/25 (4\%) of the total number of pixels, i.e., the total number of measurements is M = 442368.

\begin{figure}[htp]
\begin{center}
  \includegraphics[width=4in]{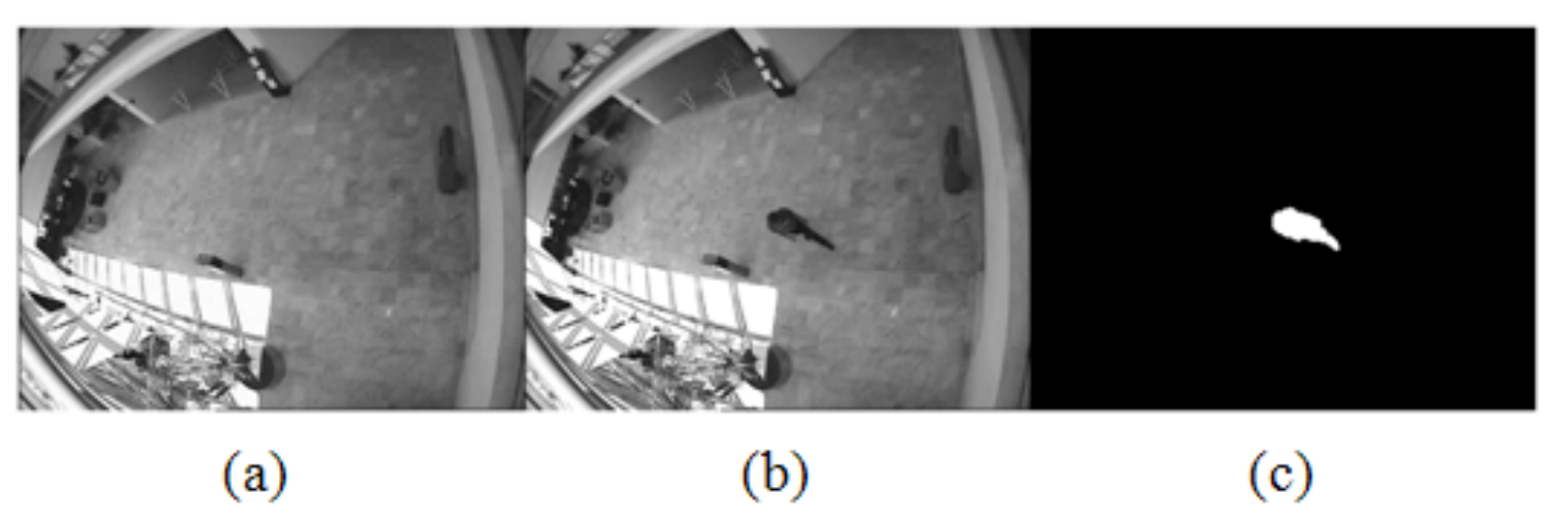}\\
  \caption{\textbf{Frame 18 of Browse2 sequence.} Total number of measurements is 4\% of the total number of pixels. (a) Reconstructed background. (b) Original frame. (c) Silhouette of the reconstructed moving objects
}\label{Browse2}
  \end{center}
\end{figure}

A typical frame, Frame 18 is shown in Figure \ref{Browse2}. The frame from the original is shown in the center (b), and the reconstructed background (the low rank component) is shown on the left, and the silhouette of the reconstructed moving objects (the sparse component) is shown in the right.

\subsection{ShopAssistant1Front}

\begin{figure}[htp]
\begin{center}
  \includegraphics[width=4in]{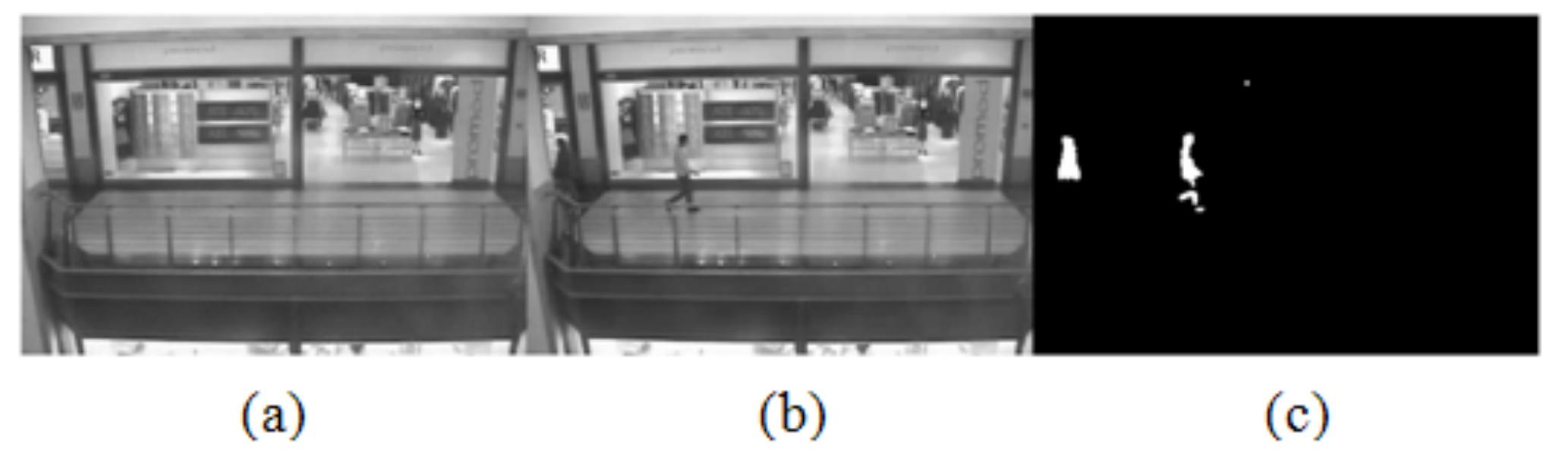}\\
  \caption{\textbf{Frame 115 of ShopAssistant1front sequence.} Total number of measurements is 4\% of the total number
of pixels. (a) Reconstructed background. (b) Original frame. (c) Silhouette of the reconstructed moving objects.} \label{Shop}
  \end{center}
\end{figure}

ShopAssistant1Front \cite{None} is a color sequence from a camera in a shopping mall. The original is an MPEG
file of resolution $384\times258$ and about 1 minute in length. We take 120 frames from the sequence, and process
only the luminance component. The total number of measurements used in the reconstruction is 4\% of the total number of the pixels. Frame 115 is shown in Figure \ref{Shop}.

It is worthwhile to note that even with only 4\% measurements, we are able to extract the people moving inside the shops both above and below the main floor. This is shown in the small white dot in the upper middle region of Figure \ref{Shop} (c), which represents shoppers in the shop above the floor behind the shelf.

\subsection{Traffic}

\begin{figure}[htp]
\begin{center}
  \includegraphics[width=4in]{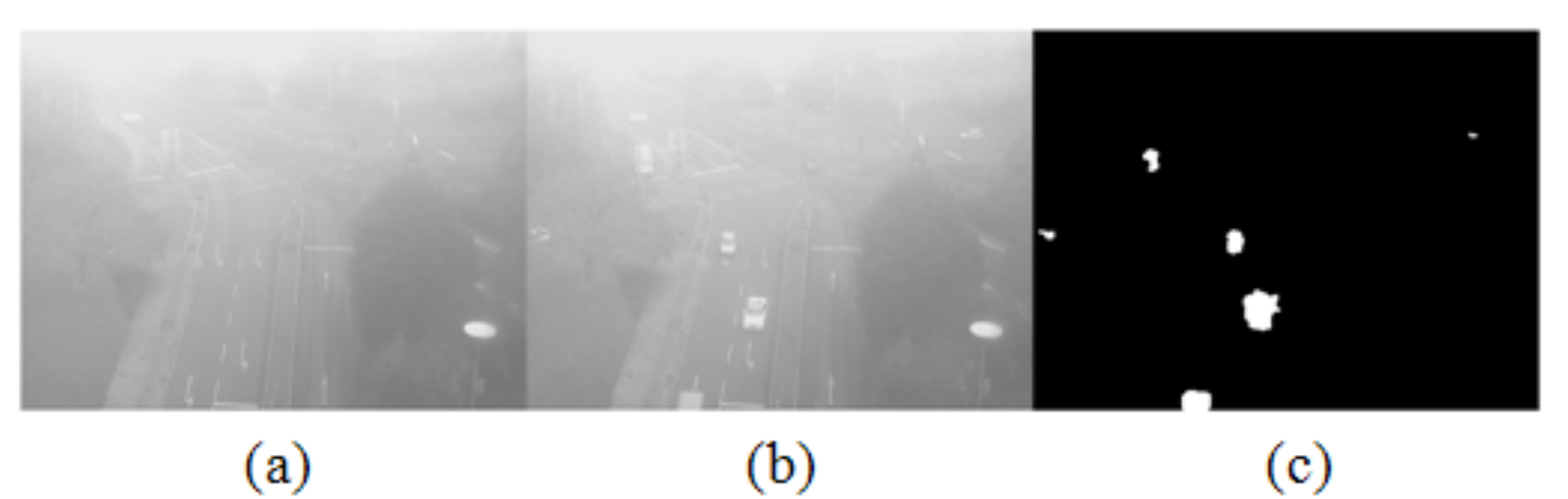}\\
  \caption{\textbf{Frame 155 of Traffic sequence.} Total number of measurements is 6.67\% of the total number of pixels.
(a) Reconstructed background. (b) Original frame. (c) Silhouette of the reconstructed moving objects} \label{traffic}
  \end{center}
\end{figure}

Traffic \cite{Mahadevan} is a black and while sequence from a traffic camera in a highway intersection. The original is a
sequence of 190 JPEG frames of resolution 378x282. The total number of measurements used in the reconstruction
is 6.67\% (1/15) of the total number of pixels. Frame 155 is shown in Figure \ref{traffic}.

As can be seen from Figure \ref{traffic}, with 6.67\% measurements, all moving vehicles are removed from the background (a).
All vehicles, except one, are detected in (c). The undetected car is in the lane going north (up), close to the intersection in the upper middle region of Figure \ref{traffic} (b). The color (intensity) of the car is very close to that of the road, and the car is indistinguishable from the noise in the reconstructed sparse component.
\subsection{Daniel\_light}
\begin{figure}[htp]
\begin{center}
  \includegraphics[width=4in]{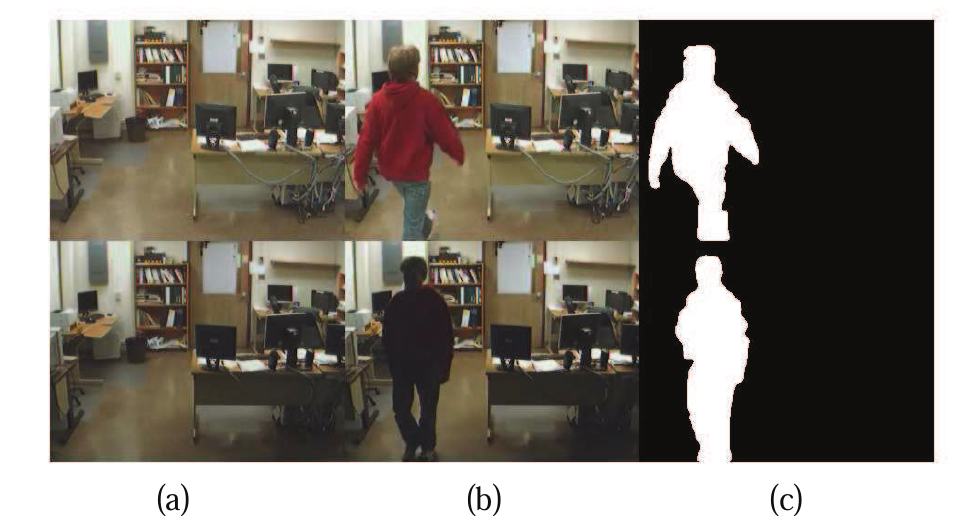}\\
  \caption{\textbf{. Frames 22 and 102 of Daniel\_light sequence.} Total number of measurements is 10\% of the total number of pixels. Top: frame 22. Bottom: frame 102. (a) Reconstructed background. (b) Original frame. (c) Silhouette of the reconstructed moving objects } \label{Daniel}
  \end{center}
\end{figure}

Daniel\_light \cite{DongY} is a color sequence from a camera monitoring an office. The original is a WMV file of resolution 320x240 and about 30 seconds in length. We take 130 frames from the sequence and process the full color with joint color components. Within the sequence, Daniel walks into the office while the light is on, turns of the light and walks out. Therefore, there is an illumination change in the sequence. Frames 22 and 102 are shown in Figure \ref{Daniel}.

Frame 22, the top row of Figure \ref{Daniel}, shows Daniel walks in while the light is on. In Frame 102, the bottom row of Figure \ref{Daniel}, Daniel walks out after turning off the light. The background Figure \ref{Daniel} (a) is well captured when light is either on or off. Note that the illumination change in this sequence is not a simple scaling of the background: only the light in front is turned off. The significance of this experiment is that in our method, the change in the illumination is not detected as part of moving objects.

\section{Conclusion}
 Low rank and sparse decomposition is an effective method for processing surveillance video when it is combined with  compressive sensing. This method is a good reconstruction method of the surveillance video because it takes advantage of the well defined low rank and sparse components in the surveillance video signal. The background subtraction and moving object extraction come from the process of reconstruction at no additional cost. We have demonstrated by experiments that moving objects can be reliably extracted by using a small amount of measurements.

The method proposed in this paper is an ``offline'' method, meaning that the processing is done after a large number of frames are acquired (using compressive measurements), and therefore, it is not done in real time. It is possible to extend the concept of this paper to ``online'', real time processing by adaptively update the low rank component. This will be investigated in details in a future paper.

\section*{Acknowledgment} The authors would like to thank Raziel Haimi-Cohen, Gang Huang, Paul Wilford, Kim Matthews, and Larry O'Gorman of Alcatel-Lucent for stimulating discussions. The authors also thank the anonymous reviewers for their valuable comments leading to the improved presentation of the paper.


\medskip
Received xxxx 20xx; revised xxxx 20xx.
\medskip

\end{document}